\pdfoutput=1
\documentclass[11pt]{article}
\usepackage[preprint]{acl}
\usepackage{times}
\usepackage{latexsym}
\usepackage[T1]{fontenc}
\usepackage[utf8]{inputenc}
\usepackage{microtype}
\usepackage{inconsolata}
\usepackage{graphicx}
\usepackage{booktabs}
\usepackage{amsmath}
\usepackage{subcaption}

\newcommand{\sat}{SAT}
\newcommand{\lap}{LAP}

\title{SAT: Structure-Aligned Tokenization with Chemical Knowledge for Explainable SMILES Language Models}

\author{
  Hyeonseok Kang \quad Goun Pyeon \quad Sangkeun Jung\thanks{Corresponding author.} \\
  Department of Computer Science and Engineering, Chungnam National University \\
  \texttt{dnfldjaak11@gmail.com}, \texttt{vusrhdns714@gmail.com}, \texttt{hugmanskj@gmail.com}
}

\begin{document}
\maketitle

\begin{abstract}
SMILES-based chemical language models perform well at property prediction, but equivalent SMILES can yield different predictions.
Stable predictions alone do not show which fragments contribute to a property or whether their contributions agree with measured changes.
We propose \textbf{Structure-Aligned Tokenization (SAT)} to combine notation invariance with fragment information, supported by \textbf{Lattice-Aligned Pretraining (LAP)} to learn from this information.
SAT defines atom representations, explanation fragments, and a removal operator from the molecular graph. LAP uses this operator to learn partial-molecule descriptors and fragment contributions.
SAT produced identical inputs across all tested notations, with no unknown tokens in any of six molecule sets.
Using the same six-layer MoLFormer architecture, SAT + LAP achieved the highest correlation with measured substitution effects on six of eight fragment-aligned tasks compared with masked language modeling baselines. It also maintained or improved on the strongest competing tokenizer's predictive performance on 11 of 18 tasks.
Gains varied by task, but the method provided invariant predictions and fragment-level evidence while retaining competitive performance on several tasks.
\end{abstract}

\section{Introduction}

Chemical language models perform well at predicting molecular properties from sequences in the Simplified Molecular Input Line Entry System (SMILES)~\citep{ross2022molformer,ahmad2022chemberta2,sultan2024transformers}.
Tokens shape how models represent molecules, motivating prior work on chemically meaningful tokens to improve prediction~\citep{li2021smilespe,ucak2023atominsmiles,noutahi2024safe}.

\begin{figure}[t!]
\centering
\includegraphics[width=\linewidth]{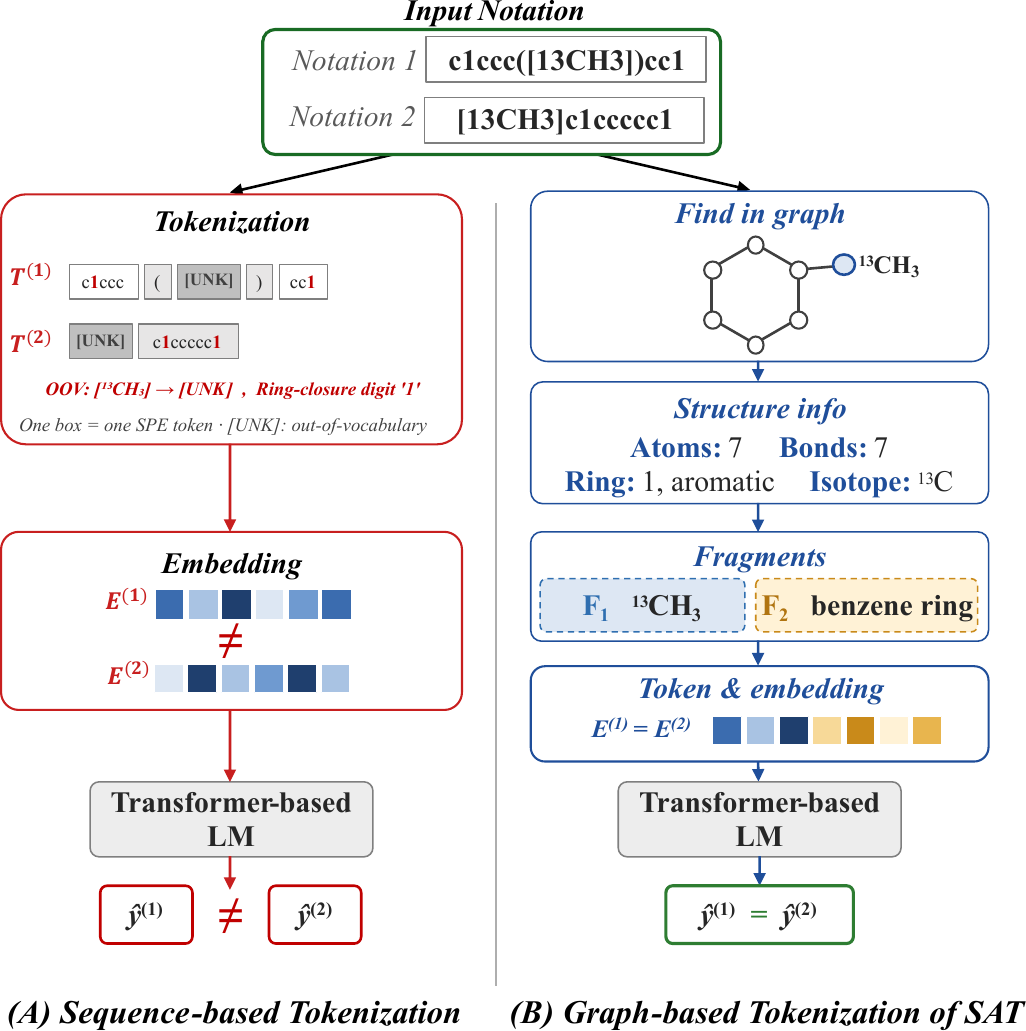}
\caption{Tokenization of two SMILES notations of one molecule. (A) Without canonicalization, SMILES Pair Encoding (SPE) can yield different token sequences $T$, embeddings $E$, and predictions $\hat y$. (B) Structure-Aligned Tokenization derives shared inputs and fragment membership from the graph, yielding identical predictions. Superscripts identify notations; $F_1,F_2$ mark fragments. LM means language model; OOV and [UNK] denote out-of-vocabulary and unknown tokens. Token splits, vocabulary gaps, and fragment boundaries are illustrative.}
\label{fig:overview}
\end{figure}

Even with chemically meaningful tokens, equivalent SMILES can yield different token sequences and predictions~\citep{ganeeva2024lost,ganeeva2025measuring}.
Canonical SMILES removes this variation by assigning one notation per molecule~\citep{oboyle2012towards}, while augmentation with multiple notations does not guarantee identical predictions~\citep{bjerrum2017smiles}.

Consistent predictions alone do not identify which fragments contribute to a property. Atom- and frequency-based token boundaries can differ from fragment boundaries, requiring a mapping from token contributions to fragments.
Substructure Mask Explanation (SME) uses prediction changes after masking~\citep{wu2023sme}, and other methods sum atom contributions within fragments~\citep{yuan2022explainable}.
These methods explain a trained model without directly teaching token representations to capture fragment contributions.
We aim to learn contributions from fragment-linked inputs and assess agreement with measured property changes~\citep{jimenezluna2020drug,wellawatte2023perspective}.

We propose \textbf{Structure-Aligned Tokenization (SAT)} to link notation-invariant atom representations to fragments (Figure~\ref{fig:overview}).
SAT derives atom representations and order from the molecular graph to give equivalent SMILES identical inputs, combining attributes such as element and charge to cover atom expressions absent from training.
It assigns atom tokens to fragments and defines how to remove them to measure contributions from prediction changes.

Masked language modeling (MLM) does not directly target fragment removal responses or contributions. \textbf{Lattice-Aligned Pretraining (LAP)} learns them by predicting partial-molecule descriptors and fragment contributions under SAT's removal rules.
An auxiliary fine-tuning objective matches summed contributions to the task prediction. We evaluate agreement with measured changes separately.

We ask three research questions (RQs):
\begin{itemize}
\item \textbf{RQ1:} Can SAT preserve inputs, predictions, and fragment contributions across notations of the same molecule while covering unseen OpenSMILES atom-attribute combinations~\citep{james2016opensmiles}?
\item \textbf{RQ2:} How well do learned fragment contributions match measured effects of fragment-aligned substitutions, and how does LAP compare with MLM?
\item \textbf{RQ3:} Can SAT + LAP provide these properties while matching the strongest competing tokenizer on each task under the same architecture and training budget?
\end{itemize}

We compare SAT + LAP with SMILES Pair Encoding (SPE), Atom-in-SMILES (AIS), Smirk, and Sequential Attachment-based Fragment Embedding (SAFE), each pretrained with token-level MLM in the same six-layer MoLFormer.
SAT gave identical token sequences across 20 notations, invariant predictions and fragment contributions on all 13 evaluated tasks, and no unknown tokens (UNK) in six molecule sets.
Canonical SMILES, AIS, and SAFE also gave invariant predictions. SAT combines invariance with atom-expression coverage and fragment information.

On fragment-aligned pairs, SAT + LAP had the highest correlation with measured substitution effects on six of eight tasks against MLM baselines and significantly outperformed SAT + MLM on four.
On Clearance, explanations followed prediction differences that poorly matched measurements (explanation correlation $0.172$, rising to $0.353$ without the auxiliary loss).
AIS with descriptor regression significantly outperformed SAT + LAP on four tasks and AIS + MLM on seven (Appendix~\ref{app:mtr}).

\begin{figure*}[t!]
\centering
\includegraphics[width=\textwidth]{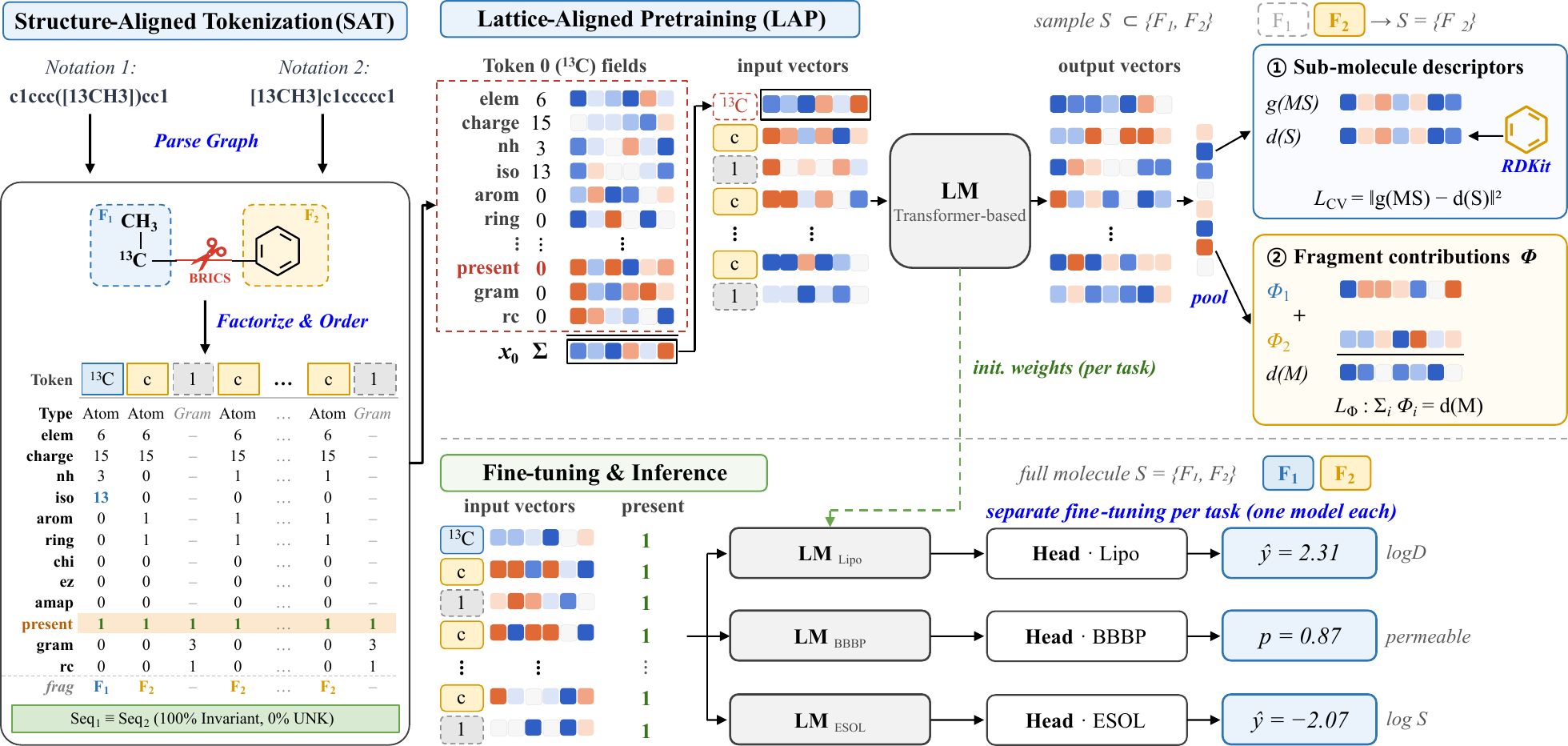}
\caption{\sat{} and \lap{} overview. \textbf{Left:} \sat{} builds 12 fields from RDKit graphs and canonical SMILES; \texttt{frag} records BRICS fragments $F_i$. \textbf{Top:} $x_0$ is the sum of field embeddings; $M_S$ sets \texttt{present}=0 outside subset $S$. The coalition-value loss $\mathcal L_{\mathrm{CV}}$ fits $g(M_S)$ to descriptors $d(S)$; $\mathcal L_{\Phi}$ supervises contributions $\Phi_i$ and their sum against whole-molecule descriptors $d(M)$. RDKit supplies the descriptor and Shapley targets. \textbf{Bottom:} LAP initializes separate models for lipophilicity (Lipo), blood--brain barrier penetration (BBBP), and estimated solubility (ESOL). Illustrative outputs: $\hat y$ predicts logD, the log distribution coefficient, or logS, the log aqueous solubility; $p$ is a class probability.}
\label{fig:method}
\end{figure*}

Against each task's strongest competing tokenizer, SAT + LAP was better on three tasks and on par on eight (11 of 18). Six of the seven tasks that failed the preservation criterion had estimated relative losses of at most $5\%$, with upper confidence bounds beyond the margin.
These results support combining invariant inputs and learned fragment contributions with competitive prediction, with gains depending on task and training objective.
The study makes three contributions:
\begin{itemize}
\item \textbf{Invariant inputs linked to chemical fragments.} SAT defines atom sequences, fragment membership, and removal rules while covering unseen atom-attribute combinations.
\item \textbf{Learning fragment responses and contributions.} LAP supervises SAT's fragment information through partial-molecule descriptors and contribution targets.
\item \textbf{Agreement with measured effects and competitive prediction.} SAT + LAP ranks highest relative to MLM baselines on six of eight fragment-aligned tasks and preserves predictive performance on 11 of 18 tasks.
\end{itemize}

\section{Related Work}

\subsection{Tokenization for Chemical Language Models}

Atom-level tokenizers split SMILES with regular expressions~\citep{schwaller2018found}, and SPE merges frequent atom-token sequences~\citep{li2021smilespe}.
AIS adds ring and neighborhood information~\citep{ucak2023atominsmiles}; SAFE uses fragment-based representations~\citep{noutahi2024safe}.
Smirk splits bracketed atoms into OpenSMILES components~\citep{wadell2026smirk}.

\subsection{Explaining and Evaluating Fragment Contributions}

SME computes contributions from prediction changes after masking substructures~\citep{wu2023sme}; fragment-level Shapley values average marginal contributions over fragment subsets~\citep{roth2026fragment}.
Substructure Explanation via Attribution Learning (SEAL) links contributions to predictions by limiting inter-fragment message passing~\citep{musial2025seal}.
These explanations can be checked against known contributions~\citep{rao2022quantitative} or measured substitution effects through matched molecular pair (MMP) analysis, which compares molecules differing in one substituent~\citep{hussain2010computationally,griffen2011matched}.

\subsection{Pretraining Objectives}

MoLFormer uses MLM~\citep{ross2022molformer}; ChemBERTa-2 also uses multi-task regression (MTR) of molecular descriptors~\citep{ahmad2022chemberta2}.
For contribution estimation, ViT-Shapley trains vision transformers on partial inputs~\citep{covert2023vitshapley}.

Our study connects invariant inputs, fragment contribution learning, and evaluation against measured changes through SAT and its supporting objective, LAP.

\section{Structure-Aligned Tokenization with Lattice-Aligned Pretraining}
\label{sec:method}

SAT derives invariant atom fields, fragments, and a removal operator from molecular graphs.
LAP learns partial-molecule descriptors and fragment contributions from these inputs. Its pretrained weights then initialize task models using whole-molecule inputs (Figure~\ref{fig:method}).

\subsection{Notation-Invariant Atom-Field Sequence}
\label{sec:sat}

Atom order and stereochemical symbols such as \texttt{@}/\texttt{@@} depend on SMILES notation~\citep{james2016opensmiles}.
SAT removes this dependence by parsing SMILES with RDKit~\citep{landrum2026rdkit} and deriving token values and order from the molecular graph.

Each atom has ten fields: atomic number (\texttt{elem}), formal charge (\texttt{charge}), hydrogen count (\texttt{nh}), isotope mass number (\texttt{iso}), aromaticity (\texttt{arom}), ring membership (\texttt{ring}), tetrahedral stereochemistry (\texttt{chi}), double-bond stereochemistry (\texttt{ez}), atom-map number (\texttt{amap}), and presence (\texttt{present}).
The charge field stores formal charge plus 15.
Under the Cahn--Ingold--Prelog (CIP) rules, \texttt{@}/\texttt{@@} become R/S labels, and \texttt{/} and \texttt{\textbackslash} become E/Z labels.
Each field has a finite set of values with learned embeddings; summing one embedding per field gives the input vector $x_0$ in Figure~\ref{fig:method}.
This factored representation~\citep{sennrich2016linguistic} covers unseen combinations of field values without UNK or vocabulary training.

Atom fields alone do not specify bonds. SAT preserves connectivity by following RDKit canonical SMILES atom order and retaining parentheses, ring closures, and bond symbols as grammar tokens.
Grammar type \texttt{gram} and ring-closure index \texttt{rc} bring the total to 12 fields; \texttt{gram} distinguishes eight symbols.
These tokens carry connectivity without pairwise biases, allowing use with linear attention~\citep{katharopoulos2020linear}.
The resulting sequence matches atom-level tokenization of canonical SMILES in length, except that separate \texttt{/} and \texttt{\textbackslash} symbols are absorbed into the \texttt{ez} field.

\subsection{Deterministic Fragments and the Present Operator}
\label{sec:fragment}

To keep explanation units fixed across notations, SAT cuts non-ring single bonds using Breaking of Retrosynthetically Interesting Chemical Substructures (BRICS) rules~\citep{degen2008brics}, excluding bonds attached to charged or hypervalent atoms.
If more than ten fragments remain, it repeatedly merges the fragment with the fewest atoms with its smallest neighbor until at most ten remain.
Canonical ranks order fragments $F_i$ and their atoms; \texttt{frag} records membership.
The fragments are notation-invariant. Because cuts lie outside rings, their graph is a tree for a connected molecule.

Let $N=\{F_1,\ldots,F_n\}$ be the fragments of molecule $M$.
For $S\subseteq N$, input $M_S$ sets \texttt{present}=0 for atoms outside $S$. For connected subsets, $v(S)$ denotes the model prediction on $M_S$ minus the all-removed baseline on $M_\emptyset$, so that $v(\emptyset)=0$.
Other fields and positions remain unchanged, including boundary hydrogen counts.
Masking instead replaces the element with \texttt{[MASK]}; hydrogen capping changes the boundary hydrogen counts.

For disconnected subsets, $v(S)$ is the sum of connected-component predictions, following a Myerson-style construction~\citep{myerson1977graphs}.
The contribution $\phi_i$ averages the prediction change from adding $F_i$ across subsets, using Shapley weights~\citep{shapley1953value,lundberg2017shap}:
\begin{equation}
\phi_i=\sum_{S\subseteq N\setminus\{F_i\}} w(S)\,\big(v(S\cup\{F_i\})-v(S)\big),
\label{eq:shapley}
\end{equation}
where $w(S)=|S|!\,(n-|S|-1)!/n!$.
The ten-fragment limit makes exact evaluation feasible, requiring at most $2^{10}=1{,}024$ model evaluations.
For a substitution between molecules $A$ and $B$, we use the difference in substituent contributions, $\hat{\Delta}=\sum\phi_B-\sum\phi_A$.

These values are notation-invariant, but invariance does not establish agreement with chemical effects. Section~\ref{sec:res-expl} tests this agreement against known contributions and measured changes.

\subsection{Lattice-Aligned Pretraining}
\label{sec:lap}

MLM reconstructs tokens~\citep{devlin2019bert}, and MTR predicts whole-molecule descriptors; neither directly supervises the removal responses $v(S)$.
Removal inputs may therefore fall outside the training distribution~\citep{hooker2019roar,frye2021manifold}.
LAP uses the same removal operator in training and explanation to learn these responses.

To obtain targets for each connected subset $S$, we cap cut bonds with hydrogen and compute 187 RDKit descriptors once before training.
Their standardized vector is $d(S)$; $d(M)=d(N)$ denotes the whole-molecule target.
Hydrogen capping is used only to compute targets; the model receives $M_S$ with the original fields and \texttt{present} flags.
We exclude Crippen logP, the log octanol--water partition coefficient, and related descriptors from pretraining targets to avoid leakage into explanation evaluation~\citep{wildman1999crippen} (Appendix~\ref{app:setup}); Appendix~\ref{app:mtr} further isolates descriptor-supervision effects via MTR baselines.

LAP uses two loss terms to learn subset responses and fragment contributions from these targets.
\begin{enumerate}
\item \textbf{Coalition-value MTR (CV-MTR):} Half the samples use $M_N$; the rest use $M_S$ for connected subsets sampled with kernel-based Shapley additive explanation (KernelSHAP) weights~\citep{lundberg2017shap,jethani2022fastshap}. The head $g$ predicts descriptors:
\begin{equation}
\mathcal{L}_{\mathrm{CV}}=\big\lVert g(M_S)-d(S)\big\rVert_2^2 .
\end{equation}
\item \textbf{Fragment Shapley regression ($\Phi$-MTR):} On whole-molecule samples, the $\Phi$ head maps each fragment's mean hidden state to a contribution vector $\Phi_i$. Equation~\ref{eq:shapley} gives its target $\phi_i^{d}$ from subset descriptors instead of model predictions. An efficiency term matches $\sum_i\Phi_i$ to $d(M)$:
\begin{equation}
\begin{aligned}
\mathcal{L}_{\Phi}={}&\tfrac{1}{n}\textstyle\sum_{i}\lVert\Phi_i-\phi_i^{d}\rVert_2^2\\
&+\lambda_{\mathrm{eff}}\big\lVert\textstyle\sum_i\Phi_i-d(M)\big\rVert_2^2 .
\end{aligned}
\end{equation}
\end{enumerate}
The total loss is $\mathcal{L}=\mathcal{L}_{\mathrm{CV}}+\lambda_{\Phi}\mathcal{L}_{\Phi}$, with $\lambda_{\Phi}=1$ and $\lambda_{\mathrm{eff}}=0.1$.
We compute $\Phi$-MTR targets $\phi_i^{d}$ only for molecules with at most eight fragments, covering 86.9\% of the corpus. This limit applies to contribution supervision; SAT still supports explanations with up to ten fragments.
Each sample requires one model evaluation.

Task fine-tuning uses whole-molecule inputs and an auxiliary loss of weight 0.1 whose fragment head maps each fragment's mean hidden state to a scalar contribution and matches their sum to the task prediction.
The auxiliary contribution head supports training; reported explanations use Shapley values computed from task predictions through Equation~\ref{eq:shapley}.
LAP supervision is limited to RDKit descriptors, and atoms switched off by \texttt{present} remain in attention.
Section~\ref{sec:ablation} tests the auxiliary loss and removal operator.

\section{Experimental Setup}
\label{sec:setup}

We compare input designs and training objectives under the same architecture and training budget (Appendix~\ref{app:setup}).

\subsection{Compared Tokenizers}

We use official implementations of Atom-level, Atom-level (canonical), SPE, AIS, Smirk, and SAFE, all pretrained with token-level MLM.
Atom-level receives randomized SMILES with probability $50\%$ during training; Atom-level (canonical) uses canonical SMILES only.
This comparison tests canonicalization; AIS and SAFE also canonicalize internally.
SAT + MLM masks fields, while SAT + LAP learns fragment responses and contributions. Comparing MLM models tests input designs; comparing SAT + MLM with SAT + LAP tests the training configurations (with the fine-tuning auxiliary loss isolated in Section~\ref{sec:ablation}).
MTR-pretrained competitors further test the role of descriptor supervision (Appendix~\ref{app:mtr}).

\subsection{Training Protocol}

All models use a six-layer MoLFormer~\citep{ross2022molformer} with linear attention, rotary positional embeddings, 768 hidden dimensions, and 12 attention heads, initialized randomly. We choose MoLFormer because linear attention admits no pairwise attention biases---strictly testing sequence-only encoding of molecular connectivity---and directly aligns with our MoleculeNet benchmarks.
The encoder architecture is fixed; input embeddings and pretraining heads vary with the tokenizer and objective.
All models are pretrained for two epochs (77,160 steps) on 9,876,638 PubChem molecules~\citep{kim2025pubchem}, excluding downstream validation and test structures.
Fine-tuning runs for 40 epochs with learning rates $\{5\times10^{-5},10^{-4}\}$; validation performance determines the rate and epoch.
Three pretraining seeds and three fine-tuning seeds give nine runs per model and task.
Optimizer and batch settings are in Appendix~\ref{app:setup}.

\subsection{Downstream Tasks}

The 18 tasks cover predictive performance and explanation validation (Table~\ref{tab:tasks} in the appendix).
We use MoleculeNet tasks from MoLFormer and ChemBERTa evaluations~\citep{wu2018moleculenet,ross2022molformer,ahmad2022chemberta2}, MoleculeACE activity-cliff tasks~\citep{vantilborg2022moleculeace}, and tasks with known fragment contributions:
\begin{itemize}
\item \textbf{Classification:} beta-secretase (BACE) inhibition~\citep{subramanian2016bace}, Toxicology in the 21st Century (Tox21)~\citep{huang2016tox21}, blood--brain barrier penetration (BBBP)~\citep{martins2012bbbp}, human immunodeficiency virus (HIV) activity~\citep{nci2004hiv}, clinical toxicity (ClinTox)~\citep{gayvert2016clintox}, and the Side Effect Resource (SIDER)~\citep{kuhn2016sider}. We use the area under the receiver operating characteristic curve (ROC-AUC; $\uparrow$).
\item \textbf{Regression:} lipophilicity (Lipo) and Clearance~\citep{hersey2015chembl}, estimated solubility (ESOL)~\citep{delaney2004esol}, FreeSolv~\citep{mobley2014freesolv}, BACE regression (BACE-R), QM8~\citep{ramakrishnan2015qm8}, and MoleculeACE CHEMBL214 (ACE-214), CHEMBL233 (ACE-233), and CHEMBL234 (ACE-234). BACE-R predicts pIC$_{50}$, the negative base-10 logarithm of the half-maximal inhibitory concentration in molar units. We use root mean squared error (RMSE; $\downarrow$); for QM8, we average standardized RMSE over its 16 targets.
\item \textbf{Explanation ground truths:} Crippen logP equals the sum of known atom contributions for 12,000 PubChem molecules. The fragment-additive task (FragAdd) sums standard-normal weights assigned to SAT fragment types, with a small noise term, on 12,000 ZINC250k molecules. The environment-additive task (EnvAdd) uses the same molecules with weights for atom environments defined by element, heavy-atom degree, aromaticity, and hydrogen count. Appendix~\ref{app:tasks} gives data sources and construction details.
\end{itemize}
MoleculeNet tasks use scaffold splits~\citep{bemis1996scaffold}, except QM8, which uses a random split; BACE-R shares the BACE classification split.
MoleculeACE retains its official training and test splits, with validation drawn from training.

\begin{figure*}[t!]
\centering
\includegraphics[width=\textwidth]{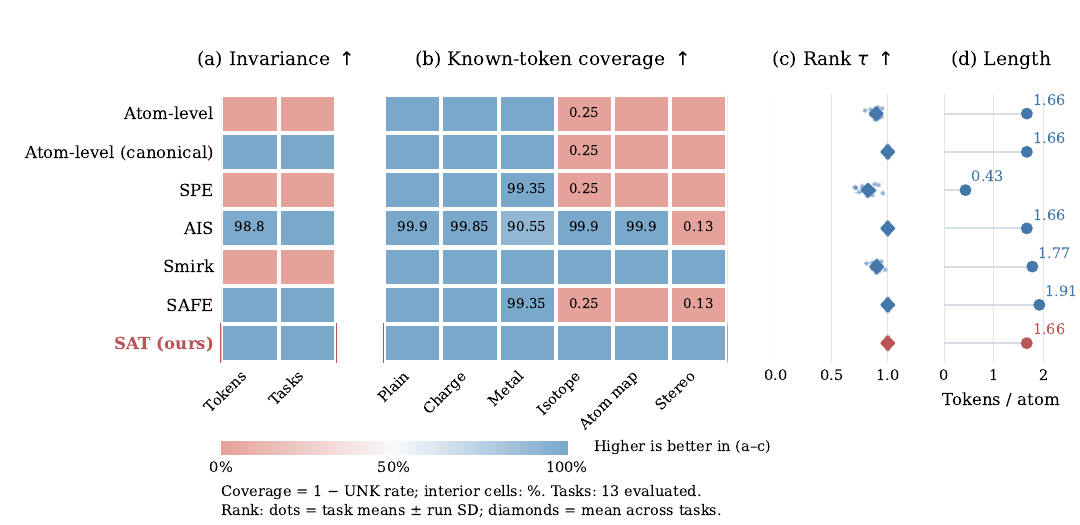}
\caption{Notation invariance and vocabulary coverage. (a) Fraction of molecules whose token sequences are identical over 20 notations (Tokens) and of tasks whose predictions and fragment Shapley values are identical over all notations (Tasks). (b) Fraction of molecules without UNK tokens in each molecule set; cell values are percentages. (c) Kendall's $\tau$ of fragment Shapley ranks across notations; dots are task means with standard deviations over runs, and diamonds are means over 13 tasks. (d) Number of tokens per atom.}
\label{fig:inv}
\end{figure*}

\subsection{Evaluation}

\textbf{Notation invariance (RQ1).}
We measure the fraction of 2,000 molecules with identical token sequences across 20 notations: 18 randomized atom orders, a Kekul\'e form, and an explicit-hydrogen form.
To test whether predictions and explanations also remain stable, we evaluate fine-tuned models on 100 molecules and ten notations per molecule across 13 tasks (all 18 tasks except Lipo, BACE, Tox21, FragAdd, and EnvAdd), measuring prediction standard deviation and Kendall's $\tau$ of fragment Shapley ranks~\citep{kendall1938tau}.
We check vocabulary coverage as the fraction of molecules without UNK tokens across six sets: general, charged, metal-containing, isotopic, atom-mapped, and stereochemical molecules.

\textbf{Agreement of fragment contributions with measurements (RQ2).}
For known contributions, we measure Pearson correlation with fragment Shapley values.
For measured effects, we extract MMPs sharing a scaffold and differing in one substituent from each task dataset~\citep{hussain2010computationally}. We retain at most 600 pairs per task whose substituent exactly matches a union of SAT fragments so that all models are evaluated on identical fragment units without splitting substituents.
We measure Spearman correlation between explanation differences $\hat{\Delta}$ and measured property differences $\Delta y=y_B-y_A$ across pairs $A,B$, with confidence intervals from scaffold bootstrapping.
The explanation operator matches pretraining: competing tokenizers and SAT + MLM use masking, while SAT + LAP uses \texttt{present}. For competing tokenizers, masking hides every token containing a removed atom.

\textbf{Preserving predictive performance (RQ3).}
Each task uses the strongest of SPE, AIS, Smirk, and SAFE by mean test performance as its reference.
The relative performance difference is the relative increase in RMSE for regression or relative decrease in ROC-AUC for classification; positive values mean worse performance.
We obtain 95\% confidence intervals from 2,000 hierarchical bootstrap samples, resampling pretraining seeds, fine-tuning seeds, then molecules~\citep{efron1993bootstrap}.
The upper bound determines the rating: \emph{better} below zero, \emph{on par} from zero through $\delta=0.077$, and \emph{worse} above $\delta$.
Before the main experiment, we set this margin to the 95th percentile of relative performance differences in a pilot varying only the pretraining seed.
Thus, a \emph{worse} rating means the performance-preservation criterion was not met, rather than necessarily showing a large mean loss.

\begin{table*}[t]
\centering
\small
\resizebox{\textwidth}{!}{%
\begin{tabular}{lccccccccccc}
\toprule
 & \multicolumn{8}{c}{MMP: Spearman($\hat{\Delta}$, $\Delta y$)} & \multicolumn{3}{c}{Synthetic: Pearson} \\
\cmidrule(lr){2-9}\cmidrule(lr){10-12}
Tokenizer & Lipo & ESOL & FreeSolv & BACE-R & Clear. & ACE-214 & ACE-233 & ACE-234 & Crippen & FragAdd & EnvAdd \\
\midrule
Atom-level & 0.699 & 0.714 & \textbf{0.833} & 0.262 & 0.284 & 0.389 & 0.224 & 0.443 & 0.939 & 0.646 & 0.924 \\
Atom-level (canonical) & \underline{0.770} & \underline{0.729} & 0.774 & 0.387 & \textbf{0.450} & \underline{0.603} & \underline{0.453} & 0.518 & 0.937 & \underline{0.667} & 0.875 \\
SPE & 0.565 & 0.540 & 0.688 & 0.120 & 0.135 & 0.347 & 0.215 & 0.355 & 0.852 & 0.506 & 0.775 \\
AIS & 0.768 & 0.520 & 0.780 & \underline{0.494} & 0.389 & 0.580 & 0.431 & 0.579 & 0.945 & 0.576 & \textbf{0.954} \\
Smirk & 0.698 & 0.594 & 0.755 & 0.273 & 0.269 & 0.360 & 0.273 & 0.398 & 0.915 & 0.474 & 0.890 \\
SAFE & 0.654 & 0.703 & 0.812 & 0.446 & 0.236 & 0.400 & 0.266 & 0.504 & 0.943 & 0.641 & 0.929 \\
\midrule
SAT + MLM & 0.702 & 0.608 & 0.756 & 0.444 & \underline{0.410} & 0.529 & 0.425 & \underline{0.603} & \underline{0.953} & 0.633 & \underline{0.949} \\
SAT + LAP (ours) & \textbf{0.864} & \textbf{0.830} & \underline{0.816} & \textbf{0.575} & 0.172 & \textbf{0.670} & \textbf{0.484} & \textbf{0.613} & \textbf{0.977} & \textbf{0.696} & 0.916 \\
\bottomrule
\end{tabular}}
\caption{Agreement of fragment explanations with ground truths ($\uparrow$). MMP columns are Spearman correlations between explanation differences and measured differences of matched molecular pairs; synthetic columns are Pearson correlations between fragment Shapley values and true contributions. Competing tokenizers and \sat{}+MLM use masking; \sat{}+\lap{} uses \texttt{present}. Clear. abbreviates Clearance. Bold denotes the best value and underlining the second best.}
\label{tab:expl}
\end{table*}

\section{Experimental Results}
\label{sec:results}

We first examine prediction invariance and agreement between fragment contributions and measured effects, then test whether predictive performance remains competitive.

\subsection{Notation Invariance and Vocabulary Coverage (RQ1)}

\textbf{Only SAT combined notation invariance with an open vocabulary among the tested methods} (Figure~\ref{fig:inv}).
SAT produced identical token sequences across all 20 notations, with unchanged predictions and fragment Shapley values on all 13 tasks.
It also produced no UNK in any of the six molecule sets.
Like the atom-level tokenizer, SAT produced 1.66 tokens per atom.

\textbf{Canonicalization also provided notation invariance.}
Atom-level (canonical), AIS, and SAFE were invariant on all 13 tasks; the latter two use internal canonicalization.
Their closed vocabularies, however, still produced UNK in some evaluation sets.
AIS produced UNK for 9.5\% of molecules in the metal set.
Conversely, Smirk had no UNK but was not notation-invariant (Shapley $\tau=0.901$).

\textbf{Reconstruction tested whether inputs retained molecular structure.}
Reconstruction from atom fields and bonds gave matching International Chemical Identifier keys (InChIKeys)~\citep{heller2015inchi} for all molecules in the five non-stereochemical sets, and for 97.9\% in the stereochemical set.

\subsection{Explanation Alignment (RQ2)}
\label{sec:res-expl}

\textbf{SAT + LAP best matched measured substitution effects on six of eight MMP tasks} (Table~\ref{tab:expl}).
Correlations were significantly higher than those of all four competing tokenizers on Lipo, ESOL, BACE-R, ACE-214, and ACE-233: scaffold-bootstrap 95\% intervals for the differences excluded zero (Table~\ref{tab:app-mmp} in the appendix).
On ACE-234, SAT + LAP significantly outperformed SPE, Smirk, and SAFE, but not AIS (difference 0.034 [$-$0.010, 0.082]).
For the same SAT input, LAP also significantly outperformed MLM on Lipo, ESOL, BACE-R, and ACE-214.
The MMP comparisons use fragment-aligned pairs, most of which include training molecules (Appendix~\ref{app:results}).

\textbf{Clearance remained a weak case.}
Its correlation was 0.172, below every model except SPE (0.135) and well below Atom-level (canonical) (0.450) and SAT + MLM (0.410).
Explanations correlated strongly with prediction differences (0.906), but those differences correlated weakly with measurements (0.197).
This suggests that inaccurate prediction differences limit agreement with measurements even when explanations closely follow the model.

The synthetic tasks test agreement where fragment contributions are known. SAT + LAP ranked highest on Crippen (0.977) and FragAdd (0.696).
On EnvAdd, however, AIS (0.954) and SAT + MLM (0.949), both explained with masking, outperformed SAT + LAP with \texttt{present} (0.916). Section~\ref{sec:ablation} examines this result.

With descriptor pretraining, AIS + MTR significantly outperformed SAT + LAP on four MMP tasks, with no significant losses, and AIS + MLM on seven (Appendix~\ref{app:mtr}).
The six-task lead is therefore limited to MLM-pretrained competitors.

\subsection{Predictive Performance Preservation (RQ3)}
\label{sec:res-perf}

\begin{table}[t]
\centering
\small
\resizebox{\columnwidth}{!}{%
\begin{tabular}{lccc}
\toprule
Tokenizer & 1st / 2nd & Better / On par / Worse & Ref. \\
\midrule
Atom-level & \underline{1} / 3 & 0 / \underline{8} / 10 & -- \\
Atom-level (canonical) & \underline{1} / 1 & 0 / \textbf{9} / 9 & -- \\
SPE & 0 / 0 & 0 / 1 / 16 & 1 \\
AIS & \underline{1} / \underline{4} & 0 / 3 / \textbf{4} & 11 \\
Smirk & \underline{1} / 1 & 0 / 3 / 12 & 3 \\
SAFE & 0 / 2 & 0 / 3 / 12 & 3 \\
\midrule
SAT + MLM & \underline{1} / \textbf{7} & 0 / \underline{8} / 10 & -- \\
SAT + LAP (ours) & \textbf{13} / 2 & \textbf{3} / \underline{8} / \underline{7} & -- \\
\bottomrule
\end{tabular}}
\caption{Performance over 18 tasks. 1st / 2nd counts first- and second-place finishes (including ties at displayed precision) among eight models. Better / On par / Worse gives counts relative to each task's strongest competing tokenizer across the $18-\text{Ref.}$ judged tasks; Ref.\ counts tasks where the model is the reference. Bold marks the best count; underlining marks the second best.}
\label{tab:perf}
\end{table}

\textbf{SAT + LAP maintained or improved predictive performance on 11 of 18 tasks} (Table~\ref{tab:perf}).
It was better on three (FragAdd, ACE-214, ACE-234), on par on eight, and worse on seven (BACE, SIDER, FreeSolv, BACE-R, Clearance, Crippen, EnvAdd).
It had the best mean on 13 tasks and the second best on two (Tables~\ref{tab:cls}, \ref{tab:reg}, and \ref{tab:app-abs}).
With the same input, SAT + MLM was on par on eight tasks, worse on ten, and better on none.

\textbf{Six of the seven tasks rated worse had estimated relative losses of at most $5\%$.}
Only FreeSolv exceeded this value (15.1\%); the others failed because their upper confidence bounds exceeded the margin.
BACE-R, Crippen, and EnvAdd had better means than the reference but failed this interval-based criterion.
Five of the seven worse tasks had at most 160 test molecules.
On FreeSolv (65 molecules) and Clearance (78), no non-reference model passed the criterion.

\subsection{Ablation Study}
\label{sec:ablation}

Table~\ref{tab:ablation} separates the effects of grammar tokens, the auxiliary loss, the pretraining objective, and the explanation operator.

\begin{table}[t]
\centering
\small
\resizebox{\linewidth}{!}{%
\begin{tabular}{lccccc}
\toprule
Variant & Perf. & MMP & Crip. & Frag. & Env. \\
\midrule
\sat{}+\lap{} (default) & 3/8/7 & -- & 0.977 & 0.696 & 0.916 \\
w/o grammar tokens & 2/9/7 & 1/2 & 0.964 & 0.650 & 0.920 \\
w/o auxiliary loss & 5/7/6 & 1/3 & 0.976 & 0.690 & 0.951 \\
\lap{} $\rightarrow$ MLM & 0/8/10 & 1/4 & 0.953 & 0.633 & 0.949 \\
\midrule
Operator: mask & -- & -- & 0.924 & 0.585 & 0.942 \\
Operator: present (joint) & -- & -- & \textbf{0.983} & \textbf{0.735} & \underline{0.977} \\
Operator: drop (joint) & -- & -- & \underline{0.982} & \underline{0.672} & \textbf{0.986} \\
\bottomrule
\end{tabular}}
\caption{Ablation of \sat{}+\lap{}. Perf.\ summarizes predictive performance as better/on-par/worse counts out of 18 tasks; MMP counts datasets (of 8) where the variant significantly outperforms/underperforms the default model. Crip., Frag., and Env. give Pearson correlations ($\uparrow$) for Crippen, FragAdd, and EnvAdd. Operator rows evaluate the same model; joint evaluates subsets jointly instead of summing connected-component predictions (default Myerson-style evaluation). The abbreviation w/o means without. Bold denotes the best value and underlining the second best among the operator rows.}
\label{tab:ablation}
\end{table}

\textbf{Removing grammar tokens} lowered correlation on Crippen (0.977 to 0.964) and FragAdd (0.696 to 0.650) and raised it on EnvAdd (0.916 to 0.920). Better/on-par/worse counts shifted from 3/8/7 to 2/9/7.

\textbf{The auxiliary loss had mixed effects.}
Without it, better/on-par/worse counts were 5/7/6 and EnvAdd correlation rose to 0.951.
MMP agreement fell on Lipo, ESOL, and ACE-214 but rose on Clearance from 0.172 to 0.353.
Its benefit is task-dependent; the main results use the configuration with this loss.

\textbf{Explanation correlations varied by operator.}
For the same trained model, joint \texttt{present} outperformed masking on all three ground truths (0.735 vs.\ 0.585 on FragAdd).
On EnvAdd, joint \texttt{drop} excluded switched-off atoms from attention, raising correlation from 0.977 to 0.986.
Joint evaluation of disconnected subsets also gave higher correlations than the default Myerson-style sum of connected-component predictions on all three tasks.

\section{Conclusion}

SAT links notation-invariant inputs to chemical fragments, and LAP learns their contributions. In our tests, SAT combined invariance with an open vocabulary.
Against MLM baselines, SAT + LAP led on six of eight fragment-aligned MMP tasks in explanation correlation with measured substitution effects. It matched or outperformed each task's strongest competing tokenizer on 11 of 18 prediction tasks.
With the same input, gains over MLM varied by task. Clearance prediction differences poorly matched measurements, and EnvAdd agreement varied by explanation operator.
Larger models and corpora remain untested.

\section*{Limitations}

\textbf{Model generalizability.} Our experiments are limited to a six-layer MoLFormer pretrained on 9.88 million molecules. Generalizability to larger models, corpora, and alternative architectures remains unverified. Furthermore, the absence of stereochemical notation in the pretraining corpus limits the learning of stereochemical representations.

\textbf{Fragment-level explanations.} Atoms excluded by the present operator remain accessible through attention, potentially affecting fragment attribution. Explanation quality also depends on predictive accuracy, fragment boundaries, and the removal operator. The advantages of SAT + LAP over MLM baselines do not consistently extend to MTR baselines.

\textbf{Evaluation scope.} MMP evaluation is restricted to fragment-aligned substitutions and predominantly includes fine-tuning molecules, limiting conclusions about unseen molecules. Explanation alignment was not directly validated on toxicity-related tasks. Additionally, small test sets on certain tasks limit the statistical reliability of performance comparisons.

\section*{Ethical Considerations}
This study focuses on chemically informed tokenization and interpretable molecular representation learning using publicly available molecular datasets. The proposed method has the potential to support transparent molecular property prediction and scientific discovery. Toxicity-related tasks are included solely as benchmarks for evaluating the generalizability of the proposed tokenization framework. As with other molecular prediction models, potential dual-use risks remain, and fragment-level explanations should be interpreted as model-derived evidence rather than experimentally established chemical effects.

\bibliography{citation}

\clearpage
\appendix

\section{Training Setup}
\label{app:setup}

Table~\ref{tab:app-setup} summarizes the hyperparameters for pretraining and fine-tuning. The pretraining corpus consists of RDKit canonical SMILES. Each MLM baseline masks its corresponding tokens, whereas SAT + MLM masks individual fields.

\textbf{Computational environment.} Models contain 22.1--25.3M parameters and were trained using NVIDIA A100 (40 GB) GPUs. Precomputing the RDKit subset descriptors and exact Shapley targets over 9.88M molecules took 2.4 hours across 192 CPU processes. Experiments were implemented with PyTorch 2.14.0 and RDKit 2026.3.6. Baseline implementations use SmilesPE 0.0.3, atomInSmiles 1.0.2, smirk 0.3.0, and safe-mol 0.2.1. Spearman correlations are computed using SciPy 1.18.1.

\textbf{Data preprocessing.} The 187 pretraining descriptors exclude MolLogP, MolMR, QED, Ipc, and the SlogP\_VSA, SMR\_VSA, BCUT2D\_LOGP, and BCUT2D\_MR families. Scaffold-based datasets are split into training, validation, and test sets at an 80/10/10 ratio by assigning scaffold groups in descending order of size. SPE merge rules (1,642) and AIS and SAFE vocabularies are constructed from 2M pretraining molecules, while Smirk uses its released vocabulary. MMPs are extracted using RDKit matched molecular pair analysis (MMPA) single cuts, retaining substituents containing at most half of the molecule's atoms.

\textbf{Pretraining.} Figure~\ref{fig:pretrain-loss} presents the pretraining loss curves across three random seeds. Loss magnitudes are not directly comparable across different prediction objectives.

\section{Downstream Tasks}
\label{app:tasks}

Table~\ref{tab:tasks} lists the six classification, nine regression, and three explanation-ground-truth tasks.

For Crippen logP, RDKit supplies labels and atom contributions for 12,000 molecules from the PubChem pretraining source.
FragAdd uses 12,000 molecules from the ZINC250k subset~\citep{gomezbombarelli2018automatic} of ZINC~\citep{irwin2012zinc}.
Each SAT fragment type receives a standard-normal weight; the label is their sum plus a small noise term.
EnvAdd uses the same molecules and assigns weights to atom environments defined by element, heavy-atom degree, aromaticity, and hydrogen count.
All three provide known fragment contributions.

Test-set sizes range from 65 molecules for FreeSolv to 4,110 for HIV.
BACE, BACE-R, ClinTox, SIDER, FreeSolv, and Clearance have at most 160 test molecules. Their wide confidence intervals can yield a \emph{worse} rating even for small mean differences (Section~\ref{sec:res-perf}).

\section{Detailed Results}
\label{app:results}

Tables~\ref{tab:cls} and \ref{tab:reg} give task-level performance; Table~\ref{tab:app-abs} reports means and standard deviations over nine runs.
The $\downarrow$ marker reflects the upper confidence bound, so even the best mean can receive a \emph{worse} rating, as on BACE-R, Crippen, and EnvAdd.

\textbf{MMP explanation comparison} (Table~\ref{tab:app-mmp}).
The fraction of cuts whose substituent exactly matched a union of SAT fragments (Aligned) ranged from 14.8\% on FreeSolv to 51.0\% on Lipo.
Most pairs include training molecules; only 5--54 pairs per task contain two test molecules.
Of 32 comparisons with the four competing tokenizers, SAT + LAP had significantly higher correlation in 24 and lower correlation in two (AIS and Smirk on Clearance).
Against SAT + MLM, its correlation was higher on Lipo, ESOL, BACE-R, and ACE-214, and lower on Clearance.

\section{Competing Tokenizers with MTR}
\label{app:mtr}

To test descriptor supervision, we replace MLM with MTR for all four competing tokenizers, retaining their token sequences, fine-tuning, and explanation procedures (Table~\ref{tab:app-mtr}).
AIS + MTR significantly outperformed SAT + LAP on BACE-R, Clearance, ACE-233, and ACE-234, with no significant losses.
It also outperformed AIS + MLM on seven tasks, excluding FreeSolv.
SAFE + MTR outperformed SAT + LAP on ACE-234; Smirk + MTR and SPE + MTR did so on Clearance.
The main explanation advantages thus apply to MLM competitors and do not consistently extend to MTR competitors.

\onecolumn
\begingroup
\captionsetup{hypcap=false}
\noindent\begin{minipage}{\textwidth}
{\centering
\small
\resizebox{\textwidth}{!}{%
\begin{tabular}{p{2.6cm}p{6.1cm}p{6.1cm}}
\toprule
Item & Pretraining & Fine-tuning \\
\midrule
Data & PubChem, 9,876,638 molecules, 2 epochs (77,160 steps) & Task training set, 40 epochs \\
Batch size & 256 & 32 (128 for training sets with $\ge$10k molecules) \\
Optimizer & AdamW, learning rate $10^{-4}$, 1,000 warm-up steps, linear decay to 10\% & AdamW, learning rate chosen from $\{5\times10^{-5}, 10^{-4}\}$ by validation performance, 5\% warm-up \\
Regularization & weight decay 0.01 & weight decay 0.01, gradient clipping 1.0 \\
Model selection & last step & best validation epoch \\
Repetitions & 3 pretraining seeds & 3 pretraining $\times$ 3 fine-tuning seeds \\
\lap{} & whole-molecule probability 0.5, $\lambda_\Phi=1$, $\lambda_{\mathrm{eff}}=0.1$, $\Phi$ targets for $\le$8 fragments & auxiliary contribution loss weight 0.1 \\
Backbone & MoLFormer linear-attention blocks: 6 layers, 768 hidden dimensions, 12 heads, feed-forward width 768, rotary positional embeddings, 32 random features & same \\
\bottomrule
\end{tabular}}
\captionof{table}{Pretraining and fine-tuning settings, shared by all models.}
\label{tab:app-setup}\par}
\end{minipage}\par

\noindent\begin{minipage}{\textwidth}
{\centering
\captionsetup{type=figure}
\begin{subfigure}[t]{0.45\textwidth}
\centering
\includegraphics[width=\linewidth]{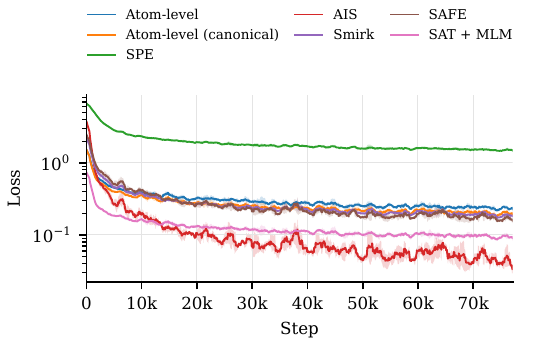}
\caption{MLM pretraining}
\end{subfigure}%
\begin{subfigure}[t]{0.45\textwidth}
\centering
\includegraphics[width=\linewidth]{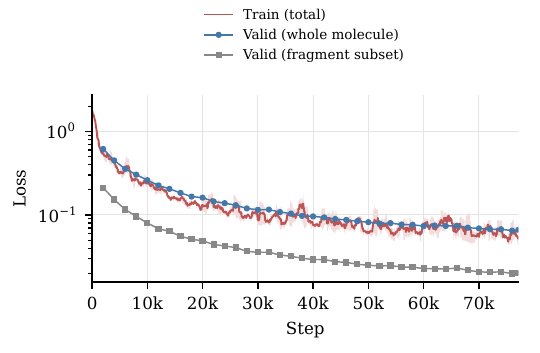}
\caption{LAP pretraining}
\end{subfigure}
\caption{Pretraining loss curves. (a) Training loss of the models pretrained with MLM. (b) Training loss of \sat{}+\lap{} and its validation loss on whole molecules and fragment subsets. Lines are means over three pretraining seeds, shaded bands span the seeds, and training losses are smoothed over 1,000 steps. Loss values are not comparable across tokenizers because their prediction targets differ.}
\label{fig:pretrain-loss}\par}
\end{minipage}\par

\noindent\begin{minipage}{\textwidth}
{\centering
\begin{minipage}{0.48\textwidth}
\small
\resizebox{\linewidth}{!}{%
\begin{tabular}{llrrl}
\toprule
Task & Type & \#Train & \#Test & Split \\
\midrule
BACE / BACE-R & Cls / Reg & 1,210 & 152 & scaffold \\
Tox21 (12) & Cls & 6,258 & 783 & scaffold \\
BBBP & Cls & 1,631 & 205 & scaffold \\
HIV & Cls & 32,896 & 4,110 & scaffold \\
ClinTox (2) & Cls & 1,184 & 142 & scaffold \\
SIDER (27) & Cls & 1,141 & 131 & scaffold \\
Lipophilicity & Reg & 3,360 & 420 & scaffold \\
ESOL & Reg & 902 & 114 & scaffold \\
FreeSolv & Reg & 513 & 65 & scaffold \\
Clearance & Reg & 669 & 78 & scaffold \\
QM8 (16) & Reg & 17,428 & 2,179 & random \\
ACE-214/233/234 & Reg & 2.4k--2.6k & 626--733 & official \\
Crippen & Reg & 10,000 & 1,000 & random \\
FragAdd / EnvAdd & Reg & 9,600 & 1,200 & random \\
\bottomrule
\end{tabular}}
\end{minipage}\par
\captionof{table}{Downstream tasks. Cls/Reg denote classification/regression; \#Train/\#Test are molecule counts. Parentheses give target counts.}
\label{tab:tasks}
\par}
\end{minipage}\par

\noindent\begin{minipage}{\textwidth}
{\centering
\small
\begin{tabular}{lcccccccc}
\toprule
Model & Lipo & ESOL & FreeSolv & BACE-R & Clear. & ACE-214 & ACE-233 & ACE-234 \\
\midrule
SPE + MTR & 0.732\textsuperscript{$\downarrow$} & 0.745\textsuperscript{$\downarrow$} & 0.722 & 0.511 & \textbf{0.610}\textsuperscript{$\uparrow$} & 0.564\textsuperscript{$\downarrow$} & 0.390\textsuperscript{$\downarrow$} & 0.548\textsuperscript{$\downarrow$} \\
AIS + MTR & \textbf{0.878} & \textbf{0.846} & 0.819 & \textbf{0.650}\textsuperscript{$\uparrow$} & \underline{0.489}\textsuperscript{$\uparrow$} & \underline{0.669} & \textbf{0.559}\textsuperscript{$\uparrow$} & \textbf{0.677}\textsuperscript{$\uparrow$} \\
Smirk + MTR & 0.808\textsuperscript{$\downarrow$} & \underline{0.835} & \underline{0.838} & \underline{0.607} & 0.384\textsuperscript{$\uparrow$} & 0.490\textsuperscript{$\downarrow$} & 0.335\textsuperscript{$\downarrow$} & 0.523\textsuperscript{$\downarrow$} \\
SAFE + MTR & 0.834\textsuperscript{$\downarrow$} & 0.817 & \textbf{0.869} & 0.563 & 0.157 & 0.614\textsuperscript{$\downarrow$} & \underline{0.513} & \underline{0.671}\textsuperscript{$\uparrow$} \\
\midrule
SAT + LAP (ours) & \underline{0.864} & 0.830 & 0.816 & 0.575 & 0.172 & \textbf{0.670} & 0.484 & 0.613 \\
\bottomrule
\end{tabular}
\captionof{table}{MMP explanation correlation, Spearman($\hat{\Delta}$, $\Delta y$) ($\uparrow$), of competing tokenizers pretrained with MTR. $\uparrow$ and $\downarrow$ mark correlations significantly higher and lower than those of \sat{}+\lap{} (scaffold-bootstrap 95\% interval). Bold denotes the best value and underlining the second best.}
\label{tab:app-mtr}\par}
\end{minipage}\par

\noindent\begin{minipage}{\textwidth}
{\centering
\small
\begin{tabular}{lcccccc}
\toprule
Tokenizer & BACE & Tox21 & BBBP & HIV & ClinTox & SIDER \\
\midrule
Atom-level & 0.829\textsuperscript{$\downarrow$} & 0.765 & 0.847 & 0.755 & 0.846\textsuperscript{$\downarrow$} & \underline{0.638} \\
Atom-level (canonical) & \underline{0.866} & 0.760 & 0.845 & 0.745 & \textbf{0.886} & 0.609\textsuperscript{$\downarrow$} \\
SPE & 0.818\textsuperscript{$\downarrow$} & 0.745 & 0.824\textsuperscript{$\downarrow$} & 0.760 & 0.779\textsuperscript{$\downarrow$} & 0.624\textsuperscript{$\downarrow$} \\
AIS & 0.862 & 0.762 & 0.848 & 0.748 & 0.824\textsuperscript{$\downarrow$} & 0.633\textsuperscript{$\downarrow$} \\
Smirk & 0.847 & 0.749 & \underline{0.863} & 0.752 & 0.843 & \textbf{0.649} \\
SAFE & 0.840 & 0.752 & 0.823\textsuperscript{$\downarrow$} & 0.732\textsuperscript{$\downarrow$} & 0.821\textsuperscript{$\downarrow$} & 0.625\textsuperscript{$\downarrow$} \\
\midrule
SAT + MLM & \textbf{0.869} & \underline{0.766} & 0.842 & \underline{0.762} & 0.848\textsuperscript{$\downarrow$} & 0.604\textsuperscript{$\downarrow$} \\
SAT + LAP (ours) & 0.826\textsuperscript{$\downarrow$} & \textbf{0.767} & \textbf{0.868} & \textbf{0.774} & \underline{0.885} & \underline{0.638}\textsuperscript{$\downarrow$} \\
\bottomrule
\end{tabular}
\captionof{table}{Test ROC-AUC ($\uparrow$) on classification tasks, averaged over nine runs. Bold denotes the best score and underlining the second best. Against the strongest competing tokenizer, $\uparrow$ marks an upper confidence bound below zero (\emph{better}); $\downarrow$ marks one above the margin $\delta$ (\emph{worse}).}
\label{tab:cls}\par}
\end{minipage}\par

\noindent\begin{minipage}{\textwidth}
{\centering
\small
\resizebox{\linewidth}{!}{%
\begin{tabular}{lccccccccc}
\toprule
Tokenizer & Lipo & ESOL & FreeSolv & BACE-R & Clear. & QM8 & ACE-214 & ACE-233 & ACE-234 \\
\midrule
Atom-level & \underline{0.704} & 0.868\textsuperscript{$\downarrow$} & 1.778\textsuperscript{$\downarrow$} & 0.933\textsuperscript{$\downarrow$} & \textbf{46.0}\textsuperscript{$\downarrow$} & 0.400 & 0.748 & 0.889\textsuperscript{$\downarrow$} & 0.788\textsuperscript{$\downarrow$} \\
Atom-level (canonical) & 0.737\textsuperscript{$\downarrow$} & 0.838\textsuperscript{$\downarrow$} & 1.815\textsuperscript{$\downarrow$} & 0.888\textsuperscript{$\downarrow$} & 47.1\textsuperscript{$\downarrow$} & 0.399 & 0.748 & 0.849 & 0.760 \\
SPE & 0.794\textsuperscript{$\downarrow$} & 0.921\textsuperscript{$\downarrow$} & 2.467\textsuperscript{$\downarrow$} & 0.939\textsuperscript{$\downarrow$} & 49.9\textsuperscript{$\downarrow$} & 0.436\textsuperscript{$\downarrow$} & 0.763\textsuperscript{$\downarrow$} & 0.889\textsuperscript{$\downarrow$} & 0.791\textsuperscript{$\downarrow$} \\
AIS & 0.710 & 0.832\textsuperscript{$\downarrow$} & \textbf{1.749} & \underline{0.884} & 47.7\textsuperscript{$\downarrow$} & \underline{0.385} & 0.744 & 0.832 & 0.756 \\
Smirk & 0.733\textsuperscript{$\downarrow$} & 0.833\textsuperscript{$\downarrow$} & 1.978\textsuperscript{$\downarrow$} & 0.903\textsuperscript{$\downarrow$} & 47.3\textsuperscript{$\downarrow$} & 0.415\textsuperscript{$\downarrow$} & 0.761\textsuperscript{$\downarrow$} & 0.931\textsuperscript{$\downarrow$} & 0.821\textsuperscript{$\downarrow$} \\
SAFE & 0.830\textsuperscript{$\downarrow$} & \underline{0.818} & 2.063\textsuperscript{$\downarrow$} & 0.958\textsuperscript{$\downarrow$} & \underline{46.1} & 0.412\textsuperscript{$\downarrow$} & 0.773\textsuperscript{$\downarrow$} & 0.916\textsuperscript{$\downarrow$} & 0.765 \\
\midrule
SAT + MLM & 0.727\textsuperscript{$\downarrow$} & 0.843\textsuperscript{$\downarrow$} & \underline{1.771}\textsuperscript{$\downarrow$} & 0.905\textsuperscript{$\downarrow$} & \underline{46.1}\textsuperscript{$\downarrow$} & 0.400 & \underline{0.729} & \underline{0.826} & \underline{0.744} \\
SAT + LAP (ours) & \textbf{0.696} & \textbf{0.760} & 2.012\textsuperscript{$\downarrow$} & \textbf{0.879}\textsuperscript{$\downarrow$} & 47.0\textsuperscript{$\downarrow$} & \textbf{0.381} & \textbf{0.707}\textsuperscript{$\uparrow$} & \textbf{0.801} & \textbf{0.702}\textsuperscript{$\uparrow$} \\
\bottomrule
\end{tabular}}
\captionof{table}{Test RMSE ($\downarrow$) on regression tasks. QM8 is the mean standardized RMSE over its targets. Notation follows Table~\ref{tab:cls}.}
\label{tab:reg}\par}
\end{minipage}\par

\noindent\begin{minipage}{\textwidth}
{\centering
\scriptsize
\resizebox{\textwidth}{!}{%
\begin{tabular}{lcccccccc}
\toprule
Task & Atom-level & Atom (canon.) & SPE & AIS & Smirk & SAFE & SAT + MLM & SAT + LAP \\
\midrule
BACE & 0.829 $\pm$ 0.017 & 0.866 $\pm$ 0.008 & 0.818 $\pm$ 0.011 & 0.862 $\pm$ 0.011 & 0.847 $\pm$ 0.018 & 0.840 $\pm$ 0.010 & 0.869 $\pm$ 0.012 & 0.826 $\pm$ 0.014 \\
Tox21 & 0.765 $\pm$ 0.011 & 0.760 $\pm$ 0.004 & 0.745 $\pm$ 0.009 & 0.762 $\pm$ 0.007 & 0.749 $\pm$ 0.005 & 0.752 $\pm$ 0.004 & 0.766 $\pm$ 0.006 & 0.767 $\pm$ 0.005 \\
BBBP & 0.847 $\pm$ 0.011 & 0.845 $\pm$ 0.017 & 0.824 $\pm$ 0.008 & 0.848 $\pm$ 0.011 & 0.863 $\pm$ 0.015 & 0.823 $\pm$ 0.015 & 0.842 $\pm$ 0.012 & 0.868 $\pm$ 0.012 \\
HIV & 0.755 $\pm$ 0.018 & 0.745 $\pm$ 0.020 & 0.760 $\pm$ 0.008 & 0.748 $\pm$ 0.015 & 0.752 $\pm$ 0.017 & 0.732 $\pm$ 0.023 & 0.762 $\pm$ 0.024 & 0.774 $\pm$ 0.009 \\
ClinTox & 0.846 $\pm$ 0.019 & 0.886 $\pm$ 0.017 & 0.779 $\pm$ 0.044 & 0.824 $\pm$ 0.037 & 0.843 $\pm$ 0.030 & 0.821 $\pm$ 0.016 & 0.848 $\pm$ 0.020 & 0.885 $\pm$ 0.027 \\
SIDER & 0.638 $\pm$ 0.017 & 0.609 $\pm$ 0.019 & 0.624 $\pm$ 0.018 & 0.633 $\pm$ 0.011 & 0.649 $\pm$ 0.018 & 0.625 $\pm$ 0.012 & 0.604 $\pm$ 0.020 & 0.638 $\pm$ 0.021 \\
Lipo & 0.704 $\pm$ 0.022 & 0.737 $\pm$ 0.015 & 0.794 $\pm$ 0.011 & 0.710 $\pm$ 0.010 & 0.733 $\pm$ 0.021 & 0.830 $\pm$ 0.018 & 0.727 $\pm$ 0.017 & 0.696 $\pm$ 0.013 \\
ESOL & 0.868 $\pm$ 0.063 & 0.838 $\pm$ 0.019 & 0.921 $\pm$ 0.044 & 0.832 $\pm$ 0.034 & 0.833 $\pm$ 0.036 & 0.818 $\pm$ 0.026 & 0.843 $\pm$ 0.018 & 0.760 $\pm$ 0.022 \\
FreeSolv & 1.778 $\pm$ 0.122 & 1.815 $\pm$ 0.062 & 2.467 $\pm$ 0.121 & 1.749 $\pm$ 0.084 & 1.978 $\pm$ 0.109 & 2.063 $\pm$ 0.149 & 1.771 $\pm$ 0.123 & 2.012 $\pm$ 0.065 \\
BACE-R & 0.933 $\pm$ 0.025 & 0.888 $\pm$ 0.024 & 0.939 $\pm$ 0.021 & 0.884 $\pm$ 0.023 & 0.903 $\pm$ 0.027 & 0.958 $\pm$ 0.033 & 0.905 $\pm$ 0.031 & 0.879 $\pm$ 0.028 \\
Clear. & 46.0 $\pm$ 1.9 & 47.1 $\pm$ 1.9 & 49.9 $\pm$ 2.8 & 47.7 $\pm$ 3.0 & 47.3 $\pm$ 2.3 & 46.1 $\pm$ 2.1 & 46.1 $\pm$ 2.7 & 47.0 $\pm$ 2.3 \\
QM8 & 0.400 $\pm$ 0.005 & 0.399 $\pm$ 0.004 & 0.436 $\pm$ 0.005 & 0.385 $\pm$ 0.005 & 0.415 $\pm$ 0.004 & 0.412 $\pm$ 0.007 & 0.400 $\pm$ 0.007 & 0.381 $\pm$ 0.004 \\
ACE-214 & 0.748 $\pm$ 0.014 & 0.748 $\pm$ 0.021 & 0.763 $\pm$ 0.009 & 0.744 $\pm$ 0.016 & 0.761 $\pm$ 0.013 & 0.773 $\pm$ 0.013 & 0.729 $\pm$ 0.019 & 0.707 $\pm$ 0.013 \\
ACE-233 & 0.889 $\pm$ 0.023 & 0.849 $\pm$ 0.024 & 0.889 $\pm$ 0.016 & 0.832 $\pm$ 0.019 & 0.931 $\pm$ 0.018 & 0.916 $\pm$ 0.025 & 0.826 $\pm$ 0.017 & 0.801 $\pm$ 0.022 \\
ACE-234 & 0.788 $\pm$ 0.017 & 0.760 $\pm$ 0.012 & 0.791 $\pm$ 0.023 & 0.756 $\pm$ 0.014 & 0.821 $\pm$ 0.016 & 0.765 $\pm$ 0.019 & 0.744 $\pm$ 0.016 & 0.702 $\pm$ 0.016 \\
Crippen & 0.333 $\pm$ 0.015 & 0.346 $\pm$ 0.011 & 0.331 $\pm$ 0.005 & 0.270 $\pm$ 0.019 & 0.373 $\pm$ 0.013 & 0.391 $\pm$ 0.025 & 0.323 $\pm$ 0.012 & 0.248 $\pm$ 0.009 \\
FragAdd & 1.252 $\pm$ 0.011 & 1.331 $\pm$ 0.019 & 1.342 $\pm$ 0.016 & 1.280 $\pm$ 0.029 & 1.320 $\pm$ 0.020 & 1.265 $\pm$ 0.020 & 1.335 $\pm$ 0.030 & 1.189 $\pm$ 0.010 \\
EnvAdd & 0.198 $\pm$ 0.011 & 0.235 $\pm$ 0.012 & 0.266 $\pm$ 0.010 & 0.168 $\pm$ 0.007 & 0.207 $\pm$ 0.010 & 0.247 $\pm$ 0.006 & 0.188 $\pm$ 0.007 & 0.165 $\pm$ 0.009 \\
\bottomrule
\end{tabular}}
\captionof{table}{Test performance on all 18 tasks (mean $\pm$ standard deviation over nine runs). ROC-AUC ($\uparrow$) for classification and RMSE ($\downarrow$) for regression.}
\label{tab:app-abs}\par}
\end{minipage}\par

\noindent\begin{minipage}{\textwidth}
{\centering
\scriptsize
\resizebox{\textwidth}{!}{%
\begin{tabular}{lccccccccc}
\toprule
Task & Mols & Aligned & Pairs & Test-only & vs SPE & vs AIS & vs Smirk & vs SAFE & vs SAT+MLM \\
\midrule
Lipo & 4,200 & 0.510 & 600 & 19 & 0.298 [0.248, 0.353] & 0.096 [0.064, 0.128] & 0.166 [0.131, 0.203] & 0.210 [0.170, 0.252] & 0.162 [0.120, 0.197] \\
ESOL & 1,117 & 0.209 & 210 & 13 & 0.290 [0.203, 0.390] & 0.310 [0.181, 0.470] & 0.236 [0.141, 0.339] & 0.128 [0.049, 0.216] & 0.222 [0.143, 0.324] \\
FreeSolv & 642 & 0.148 & 66 & 5 & 0.129 [0.020, 0.220] & 0.036 [$-$0.026, 0.127] & 0.061 [$-$0.055, 0.180] & 0.005 [$-$0.098, 0.087] & 0.061 [$-$0.033, 0.181] \\
BACE-R & 1,513 & 0.360 & 600 & 32 & 0.455 [0.348, 0.557] & 0.081 [0.013, 0.151] & 0.302 [0.228, 0.376] & 0.129 [0.064, 0.196] & 0.131 [0.074, 0.178] \\
Clear. & 830 & 0.470 & 568 & 6 & 0.037 [$-$0.052, 0.118] & $-$0.218 [$-$0.317, $-$0.124] & $-$0.097 [$-$0.189, $-$0.014] & $-$0.065 [$-$0.144, 0.011] & $-$0.238 [$-$0.326, $-$0.142] \\
ACE-214 & 3,317 & 0.508 & 600 & 54 & 0.324 [0.262, 0.386] & 0.090 [0.051, 0.131] & 0.310 [0.245, 0.374] & 0.270 [0.219, 0.323] & 0.141 [0.100, 0.178] \\
ACE-233 & 3,128 & 0.466 & 600 & 51 & 0.269 [0.177, 0.360] & 0.053 [0.001, 0.107] & 0.211 [0.133, 0.283] & 0.218 [0.146, 0.292] & 0.059 [$-$0.002, 0.110] \\
ACE-234 & 3,657 & 0.505 & 600 & 46 & 0.258 [0.186, 0.331] & 0.034 [$-$0.010, 0.082] & 0.215 [0.150, 0.275] & 0.109 [0.061, 0.166] & 0.010 [$-$0.043, 0.063] \\
\bottomrule
\end{tabular}}
\captionof{table}{MMP sets and differences in explanation correlation (\sat{}+\lap{} minus each model) with scaffold-bootstrap 95\% intervals. Mols counts molecules; Pairs counts evaluated pairs; Test-only counts pairs with both molecules in the test set. Aligned is the fraction of cuts whose substituent exactly matches a union of \sat{} fragments.}
\label{tab:app-mmp}\par}
\end{minipage}\par
\endgroup

\end{document}